\documentclass{article}

\PassOptionsToPackage{numbers, compress}{natbib}

\usepackage[preprint]{neurips_2022}

\usepackage[utf8]{inputenc} 
\usepackage[T1]{fontenc}    
\usepackage{hyperref}       
\usepackage{url}            
\usepackage{booktabs}       
\usepackage{amsfonts}       
\usepackage{nicefrac}       
\usepackage{microtype}      
\usepackage{xcolor}         

\usepackage{multirow}
\usepackage[ruled,linesnumbered]{algorithm2e}
\usepackage{graphicx}
\usepackage{amsmath}
\newcommand{\reffig}[1]{Figure \ref{#1}}

\title{Local Label Point Correction for Edge Detection of Overlapping Cervical Cells}

\author{%
Jiawei Liu\,$^{1,2,3}$, Huijie Fan\,$^{1,2}$, Qiang Wang\,$^{4,1}$, Wentao Li\,$^{1,2}$, Yandong Tang\,$^{1,2}$,\\ \textbf{Danbo Wang}\,$^{5}$, \textbf{Mingyi Zhou}\,$^{5}$ and \textbf{Li Chen}\,$^{6}$
}

\begin{document}

\maketitle

\begin{abstract}

  Accurate labeling is essential for supervised deep learning methods.
  However, it is almost impossible to accurately and manually annotate thousands of images, which results in many labeling errors for most datasets.
  We proposes a local label point correction (LLPC) method to improve annotation quality for edge detection and image segmentation tasks.
  Our algorithm contains three steps: gradient-guided point correction, point interpolation and local point smoothing.
  We correct the labels of object contours by moving the annotated points to the pixel gradient peaks.
  This can improve the edge localization accuracy, but it also causes unsmooth contours due to the interference of image noise.
  Therefore, we design a point smoothing method based on local linear fitting to smooth the corrected edge.
  To verify the effectiveness of our LLPC, we construct a largest overlapping cervical cell edge detection dataset
  (CCEDD) with higher precision label corrected by our label correction method.
  Our LLPC only needs to set three parameters, but yields 30-40$\%$ average precision improvement on multiple networks.
  The qualitative and quantitative experimental results
  show that our LLPC can improve the quality of manual
  labels and the accuracy of overlapping cell edge detection.
  We hope that our study will give a strong boost to the development of the label correction for edge detection and image segmentation.
  We will release the dataset and code at
  \url{https://github.com/nachifur/LLPC}.
\end{abstract}

\renewcommand{\thefootnote}{*}

Corresponding author\protect\footnotemark[1]: Huijie Fan, Danbo Wang

\footnotetext[1]{Email: Huijie Fan: fanhuijie@sia.cn, Danbo Wang: wangdanbo@cancerhosp-ln-cmu.com\\
$^{1}$State Key Laboratory of Robotics, Shenyang Institute of Automation, Chinese Academy of Sciences, Shenyang, China\\
$^{2}$Institutes for Robotics and Intelligent Manufacturing, Chinese Academy of Sciences, Shenyang, China\\
$^{3}$University of Chinese Academy of Sciences, Beijing, China\\
$^{4}$Key Laboratory of Manufacturing Industrial Integrated, Shenyang University, Shenyang, China\\
$^{5}$Department of Gynecology, Cancer Hospital of China Medical University, Liaoning Cancer Hospital $\&$ Institute, Shenyang, China\\
$^{6}$Department of Pathology, Cancer Hospital of China Medical University, Liaoning Cancer Hospital $\&$ Institute, Shenyang, China}

\renewcommand{\thefootnote}{\fnsymbol{footnote}}


\section{Introduction}
Medical image datasets are generally annotated by
professional physicians~\cite{Demner-Fushman2016,Almazroa2017,Johnson2019,Zhang2019,Ma2021,Lin2021,Wei2021}.
To construct an annotated dataset for edge detection or image
segmentation tasks, annotators often need to annotate points
and connect them into an object outline.
In the manual labeling process, it is difficult to control label accuracy due to human error.
Northcutt \textit{et al.}
found that label errors are numerous and universal: the average
error rate in 10 datasets is 3.4$\%$~\cite{northcutt2021pervasive}.
These wrong labels seriously affect the accuracy of model evaluation and destabilize benchmarks, which will ultimately spill over model selection and deployment.
For example, the deployed model in learning-based computer-aided diagnosis~\cite{saha2019srm,song2019joint,wan2019accurate,song2020constrained,zhang2020polar} is selected from many candidate models based
on evaluation accuracy, which means that inaccurate annotations may ultimately affect accurate diagnosis.
To mitigate labeling errors, an image is often annotated by multiple
annotators~\cite{Almazroa2017,Zhang2019,arbelaez2010contour}, which generates multiple labels for
one image.
However, even if the annotation standard is unified, differences
between different annotators are inevitable.
Another way is to correct the labels manually~\cite{Ma2021}.
In fact, multi-person annotation and manual label correction
are time-consuming and labor-intensive. Therefore, it is of
great value to develop label correction methods based on
manual annotation for supervised deep learning methods.

\begin{figure*}[t]
  \begin{center}
    \includegraphics[width=1\linewidth]{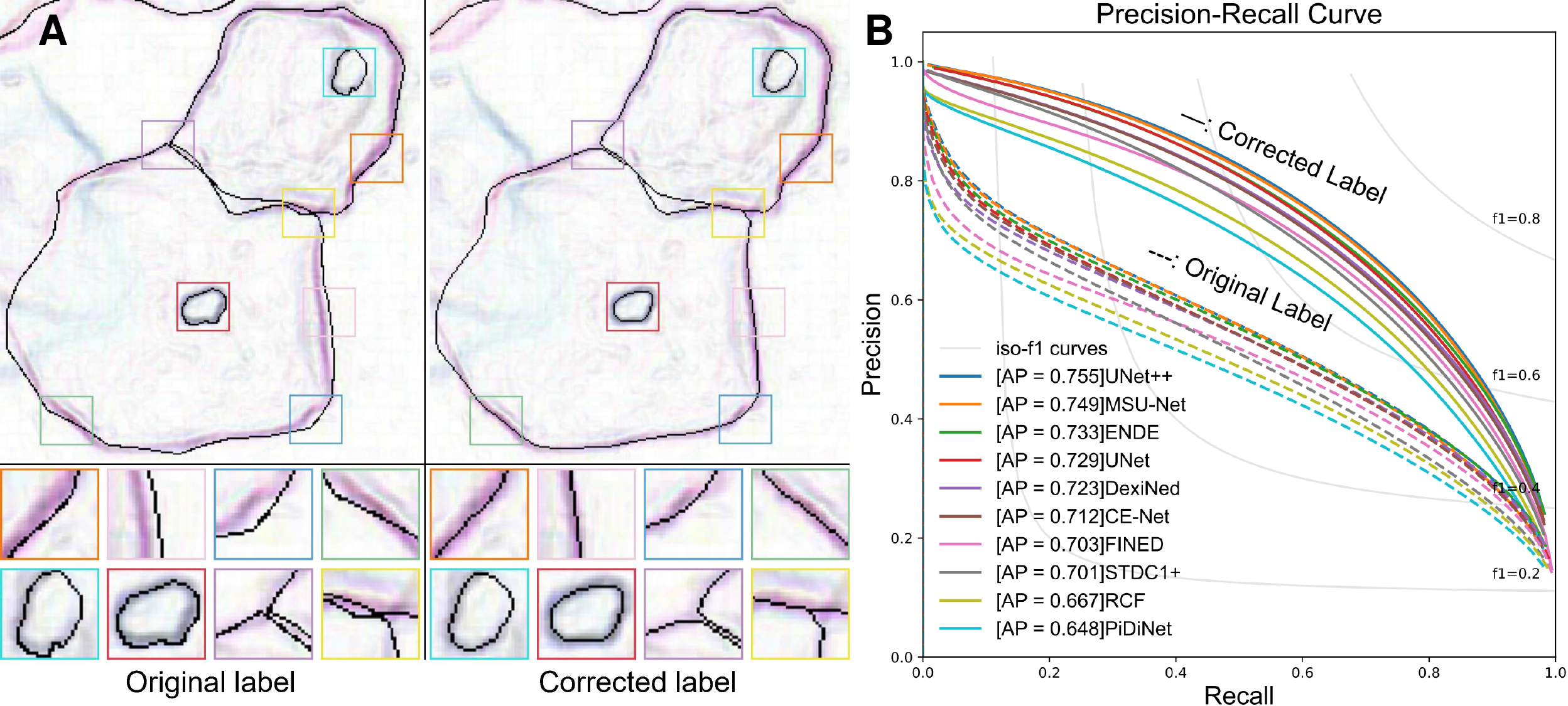}
  \end{center}
  \caption{{\bf (A)} Visual comparison of the original label and our corrected label. Our LLPC can improve the edge positioning accuracy and generate more accurate edge labels.
      {\bf (B)} Precision-Recall curves of edge detection methods on
    our CCEDD dataset. The average precision (AP) is significantly improved over multiple baseline models by using our corrected labels.}
  \label{fig1}
\end{figure*}

Most label correction works are focused on weak supervision~\cite{zheng2021meta}, semi-supervision~\cite{Li2020}, crowdsourced labeling~\cite{bhadra2015correction,nicholson2016label},
classification~\cite{nicholson2015label,kremer2018robust,guo2019lcc,Liu2020,wang2021proselflc,Li2022} and natural
language processing~\cite{zhu2019dynamic}. However, label correction in these
tasks is completely different from correcting object contours.
To automatically correct edge labels, we propose a local label point correction method
for edge detection and image segmentation.
Our method contains three steps: gradient-guided point correction, point interpolation and local point smoothing.
We correct the annotation of the object contours by moving label points to the pixel gradient peaks and smoothing the edges formed by these points.
To verify the effectiveness of our label correction method, we construct a cervical
cell edge detection dataset.
Experiments with multiple
state-of-the-art deep learning models on the CCEDD show that our LLPC can greatly improve the quality of manual
annotation and the accuracy of overlapping cell edge detection, as shown in~\textbf{\reffig{fig1}}. Our unique contributions are
summarized as follows:

\begin{itemize}
  \item We are the first to propose a label correction
        method based on annotation points for edge detection and image segmentation.
        By correcting the position of these label points, our label correction
        method can generate higher-quality label, which contributes 30-40$\%$ AP improvement on multiple baseline models.
  \item We construct a largest publicly cervical cell edge detection dataset based on our LLPC.
        Our dataset is ten times larger than the previous datasets, which greatly facilitates the development of overlapping cell edge detection.
  \item We present the first publicly available label correction benchmark for improving contour annotation.
        Our study serves as a potential catalyst to promote label correction research and further paves the way to construct accurately annotated datasets for edge detection and image segmentation.
\end{itemize}

\section{RELATED WORK}
\textbf{Label Correction.}
Deep learning is developing rapidly with the help of big computing~\cite{Jouppi2017} and big data~\cite{ImageNet,zhou2017places,Sun2017}.
Some works~\cite{radford2019language,Raffel2020,Brown2020}
focus on feeding larger models with more data for better performance and generalization,
while others design task-specific model structures and loss functions~\cite{hu2019direction,Huang2021,Zhao2022} to improve performance on a fixed dataset.
Recently, data itself has received a lot of attention.
Andrew Ng \textit{et al.}~\cite{ng2021DataCentric} led the data revolution of deep learning
and successfully organized the first ”Data-Centric AI” competition.
The competition aims to improve data quality and develop data optimization pipelines,
such as label correction, data synthesis and data augmentation~\cite{Motamedi2021}.
Competitors mine data potential instead of optimizing model structure to improve performance.
Northcutt \textit{et al.}~\cite{northcutt2021pervasive} found that if
the error rate of test labels only increases by 6$\%$, ResNet18 outperforms ResNet-50 on ImageNet~\cite{ImageNet}.
To improve data quality and accurately evaluate models, there is an urgent need to develop label correction algorithms.
In weak supervision and semi-supervision~\cite{zheng2021meta,Li2020}, pseudo
label correction is usually implemented due to the lack of supervision from real labels.
Zheng \textit{et al.}~\cite{zheng2021meta} correct the noisy labels by using a meta network for image recognition and text classification.
For supervised learning, bad data can be discarded by data preprocessing, but bad labels seem inevitable in large-scale datasets.
In crowdsourcing~\cite{bhadra2015correction,nicholson2016label}, an image is annotated by multiple people
to improve the accuracy of classification task~\cite{nicholson2015label,kremer2018robust,guo2019lcc}.
Guo \textit{et al.}~\cite{guo2019lcc} trained a model by using a small amount of data and
design a label completion method to generate labels (negative or positive) for the mostly unlabeled data.
However, label correction
in these tasks is significantly different from correcting object
contours. In this paper, to eliminate edge location errors and
inter-annotator differences in manual annotation, we propose
an label correction method based on annotation points for edge detection and
image segmentation.
Besides, we compare our LLPC with conditional random fields (CRF)~\cite{sutton2012introduction},
which is popular as post-processing for other segmentation methods~\cite{chen2017deeplab,Sun2020,Lu2021,Fan2021,Ma2022,Zhang2022}.
Dense CRF~\cite{Krahenbuhl2011} improves the labeling accuracy by optimizing energy function based on coarse segmentation images,
while our LLPC is a label correction method based on annotation points,
which are two different technical routes of label correction for image segmentation.
More discussion in Section~\ref{section5c_crf}.

\textbf{Cervical Cell Dataset.}
Currently, cervical cell datasets include ISBI 2015 challenge
dataset~\cite{lu2015improved}, Shenzhen University dataset~\cite{song2016accurate} and
Beihang University dataset~\cite{wan2019accurate}.
Supervised deep learning based methods require large amounts of data with accurate annotations.
However, the only public ISBI dataset~\cite{lu2015improved} has a small amount of data and simple image
types, which are difficult to train deep neural networks.
In this paper, we construct a largest high-accuracy cervical
cell edge detection dataset based on our label
correction method.
Our CCEDD contains overlapping cervical cell masses in a variety of complex backgrounds and high-precision corrected labels,
which are sufficient in quantity and richness to train various deep learning models.

\section{Label Correction}
\begin{figure*}[t]
  \begin{center}
    \includegraphics[width=1\linewidth]{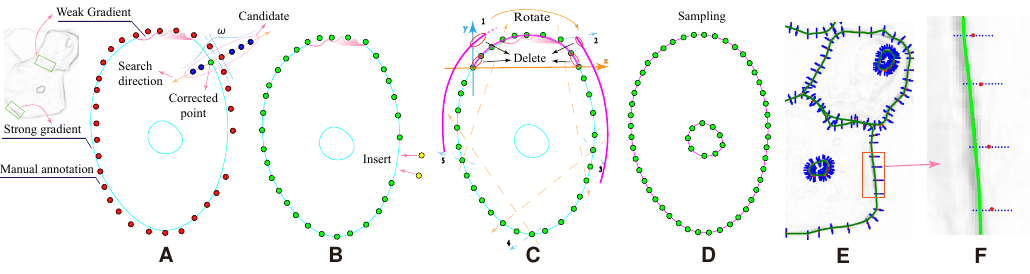}
  \end{center}
  \caption{The workflow of our LLPC algorithm. {\bf (A)} Gradient-guided point correction (the red points → the green
  points); {\bf (B)} Insert points at large intervals; {\bf (C)} Piecewise curve fitting (the purple
  curve); {\bf (D)} Curve sampling; {\bf (E)} The gradient image with the corrected edge label (the green edges); {\bf (F)} Magnification of the
  gradient image. The whole label correction process is to generate the corrected edge (green edges) from original label points (red points) in~\textbf{(F)}.}
  \label{fig2}
\end{figure*}

Our LLPC contains three steps: gradient-guided point correction (GPC), point interpolation (PI) and local point smoothing (LPS).
$I(x,y)$ is a cervical cell image and $g(x,y)$
is the gradient image of $I(x,y)$ after Gaussian smoothing.
$x_s^i$ is an original label point of $I(x,y)$.
First, we correct the points $x_s^i$ to the nearest gradient peak on $g(x,y)$,
as shown in~\textbf{\reffig{fig2}A}, i.e., $\left \{x_{s}^{i}\right \} \to \left \{x_{c}^{i}\right \}$. $i\in\left \{1,2,\dots ,n_s \right \}.$
Second, we insert more points in large gaps, as shown in~\textbf{\reffig{fig2}B},
i.e., $\left \{x_{c}^{i}\right \}\ \to \left \{x_{I}^{j}\right \}$. $j\in\left \{1,2,\dots ,n_I \right \}.$
$n_s$ and $n_I$ are the
number of points before and after interpolation, respectively.
Third, we divide the point set $\left \{x_{I}^{j}\right \}$
into $n_c$ groups. Each group
of points is expressed as $\Phi _{k}$. We fit a curve $C_k$ on $\Phi _{k}$. $k\in\left \{1,2,\dots ,n_c \right \}$. All
curves $\left \{C_{k}\right \}$ are merged into a closed curve $C_c$, as shown in~\textbf{\reffig{fig2}C}.
Finally, we sample $C_c$ to obtain discrete edges $C_d$, as shown in~\textbf{\reffig{fig2}D}.
In fact, the closed discrete edges generated by multiple curves fusion are not smooth at the stitching nodes.
Therefore, we propose a local point smoothing method without curves splicing and sampling in Section~\ref{section3c}.

\subsection{Gradient-guided Point Correction} \label{section3a}
Although the annotations of cervical cell images are provided by professional cytologists, due to human error, the label points usually deviate
from the pixel gradient peaks. To solve this problem, we design
a gradient-guided point correction (GPC) method based on gradient guidance.
We correct the label points only in the strong gradient
region to eliminate human error, while preserving the original label points
in the weak gradient region to retain the correct high-level semantics in human annotations.
Our point correction consists of three steps
as follows:
\begin{enumerate}
  \item Determine whether the position of each label point is in
        strong gradient regions.
  \item Select a set of candidate points for a label point.
  \item Move the label point to the position of the point with
        the largest gradient value among these candidate points.
\end{enumerate}

The processing object of our LLPC is a
set of label points ($\left \{x_{s}^{i}\right \}$) corresponding to a closed contour.
For an original label point $x_s^{i}$, we select candidate points
along the normal direction of label edge, as shown in \textbf{\reffig{fig2}A}.
These points constitute a candidate point set $\Omega_{x_s^i}$, and $x_{max}^i$
is the point with the largest gradient in $\Omega_{x_s^i}$. We move $x_{s}^{i}$ to
the position of $x_{max}^i$ to obtain the corrected label point $x_{c}^i$.

\begin{align}
  \label{eq1}
  x_c^i = \left\{\begin{matrix}
     & x_{max}^i & if\,\Delta >0 \\
     & x_s^i     & otherwise
  \end{matrix}\right.
\end{align}
where
\begin{equation}
  \label{eq2}
  \Delta =\left | max(\omega _{j} \cdot g(x_{s_j}^i))-min(\omega _{j} \cdot g(x_{s_j}^i)) \right |-\lambda_t \cdot max(\omega _{j}).
\end{equation}
$x_{s_j}^i$ is a candidate point in $\Omega_{x_s^i}$. We judge whether a point $x_{s}^i$
is in strong gradient regions through $\Delta$. If $\Delta > 0$, the point
will be corrected; otherwise, it will not be moved. In this
way, when the radius ($r$) of $\Omega_{x_s^i}$
is larger, our method can
correct larger annotation errors. However, this will increase
the correction error of label points due to image noise and
interference from adjacent edges. To balance the contradiction,
the gradient value of the candidate point $g(x_{s_j}^i)$ is weighted
by $\omega_{j}$, which allows setting a larger radius to correct
larger annotation errors. We compute the weight as
\begin{equation}
  \label{eq3}
  \omega_{j}=K(\left \| x_{s_j}^i-x_{s}^i \right \|_{2},h_1),
\end{equation}
where
\begin{equation}
  K(x,\,h )=\kappa(x/h)/h.
\end{equation}
$K(x,h)$ is a weighted kernel function with bandwidth $h$.
$\kappa(x)$ is a Gaussian function with zero mean and one variance.
After point correction,
$\left \{x_{s}^{i}\right \} \to \left \{x_{c}^{i}\right \}$.

\subsection{Piecewise Curve Fitting} \label{section3b}
The edge generated directly from the point set $\left \{x_{c}^{i}\right \}$
is not smooth due to the errors in point correction
process (see Section~\ref{section5c}). To eliminate the errors, we fit multiple curve segments
and stitch them together. In the annotation process of manually drawing
cell contours, the annotators perform dense point annotations near
large curvatures and sparse annotations near small curvatures
to accurately and quickly outline cell contours. Since the
existence of large intervals is not conducive to curve fitting,
we perform linear point interpolation (PI) on these intervals before curve fitting.

\begin{figure}[t]
  \centering
  \includegraphics[scale=0.8]{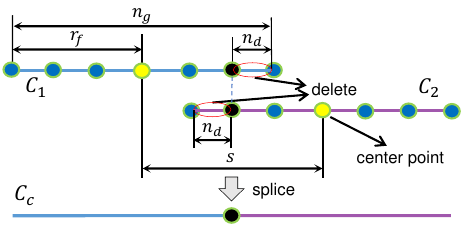}
  \caption{Merge multiple curves ($C_1$ and $C_2$) into one curve($C_c$).}
  \label{fig3}
\end{figure}

\textbf{Point Interpolation.} The sparse label point pairs can be
represented as,
\begin{equation}
  \left \{(x_{c}^{i}, x_{c}^{i+1}) |\left \|x_{c}^{i}-x_{c}^{i+1}  \right \|_2>2\cdot gap \right \},
\end{equation}
where $i=0,1\dots n_s-1$. Then, we insert points between the sparse points
pairs to satisfy
\begin{equation}
  \label{eqpi}
  \left \|x_{I}^{j}-x_{I}^{j+1}  \right \|_2< gap,
\end{equation}
as shown in \textbf{\reffig{fig2}B}.
$j=0,1\dots n_I-1$.
$n_s$ and $n_I$ are the number of points before and after
interpolation, respectively.
$gap$ is the maximum interval between adjacent point pair.
After interpolation, $\left \{x_{c}^{i}\right \}\ \to \left \{x_{I}^{j}\right \}$.

\textbf{Curve Fitting.}
We divide $\left \{x_{I}^{j}\right \}$ into $n_c$ groups.
Each group is expressed as
$\Phi _{k}=\left \{x_{I}^{1+k \cdot s},x_{I}^{2+k \cdot s}, \dots ,x_{I}^{n_g+k \cdot s} \right \}$.
$k=0,1\dots n_c-1$.
$n_c=\left \lceil n_I/s \right \rceil.$
As shown in \textbf{\reffig{fig3}}, $s=2(r_f-n_d)$ is the interval between the center points of each group;
$r_f=\left \lfloor (n_g-1)/2 \right \rfloor $ is the group radius;
$n_g$ is the number of points in the group.
To reduce the fitting error at both ends of the curve,
there is overlap between adjacent curves. The
overlapping length is $2n_d$.  To fit a curve on $\Phi _{k}$, we create
a new coordinate system, as shown in \textbf{\reffig{fig2}C}.
The x-axis
passes through the $x_I^{1+k \cdot s}$ point and the $x_I^{n_g+k \cdot s}$ point.
The point set in the new coordinate system is $\Phi _{k}^r$.
We obtain a curve $C_k$ by local linear fitting~\cite{McCrary2008} on $\Phi _{k}^r$.
This is equivalent
to solving the following problem at the target point $x_t=(x,y)$ on the
curve $C_k$.
\begin{equation}
  \label{eq6}
  \underset{\beta _0(x),\,\beta _1(x)}{min} \sum_{j=1+k \cdot s}^{n_g+k \cdot s} \omega _j(x)(y_j-\beta _0(x)-\beta _1(x)\cdot x_j )
\end{equation}
$\beta _0(x)$ and $\beta _1(x)$ are the curve parameter at the point $x_t$.
$(x_j , y_j)$ denotes the coordinates of point $x_I^j$ in $\Phi _{k}^r$.
The weight function is
\begin{equation}
  \omega _j(x)=K(\left \| x-x_j \right \|_{2},h_2)/\sum_{m=1+k \cdot s}^{n_g+k \cdot s} K(\left \| x-x_m \right \|_{2},h_2).
\end{equation}
If the distance between the point $x_I^j$ and the target point $x_t$ is larger, the weight $\omega _j(x)$ will be smaller.
The matrix representation of the above parameter solution is
\begin{equation}
  \beta =(X^T\omega X)^{-1}X^T\omega Y,
\end{equation}
where
$X=\begin{bmatrix}
    1      & x_{1+k \cdot s}   \\
    1      & x_{2+k \cdot s}   \\
    \vdots & \vdots            \\
    1      & x_{n_g+k \cdot s}
  \end{bmatrix}$,
$Y=\begin{bmatrix}
    y_{1+k \cdot s} \\
    y_{2+k \cdot s} \\
    \vdots          \\
    y_{n_g+k \cdot s}
  \end{bmatrix}$,
$\beta=\begin{bmatrix}
    \beta _0(x) \\
    \beta _1(x)
  \end{bmatrix}$,
$\omega=\begin{bmatrix}
    \omega_{1+k \cdot s}(x) &                         &        &                           \\
                            & \omega_{2+k \cdot s}(x) &        &                           \\
                            &                         & \ddots &                           \\
                            &                         &        & \omega_{n_g+k \cdot s}(x)
  \end{bmatrix}$.
The matrix $\omega$ is zero except for the diagonal. Each $\Phi _{k}^r$ corresponds to a
curve $C_k$. We stitch $n_c$ curves into a closed curve $C_c$, as shown in \textbf{\reffig{fig2}C} and \textbf{\reffig{fig3}}. Then,
we sample on the interval $\left [ x_I^{1+k \cdot s+n_d},x_I^{n_g+k \cdot s-n_d}  \right ]$
as shown in \textbf{\reffig{fig2}D}. We convert the coordinates of these sampling points to the original image coordinate system. Finally,
we can obtain a discrete edge $C_d$, as shown in \textbf{\reffig{fig2}E, F}.

\begin{figure}
  \centering
  \includegraphics[width=0.6\linewidth]{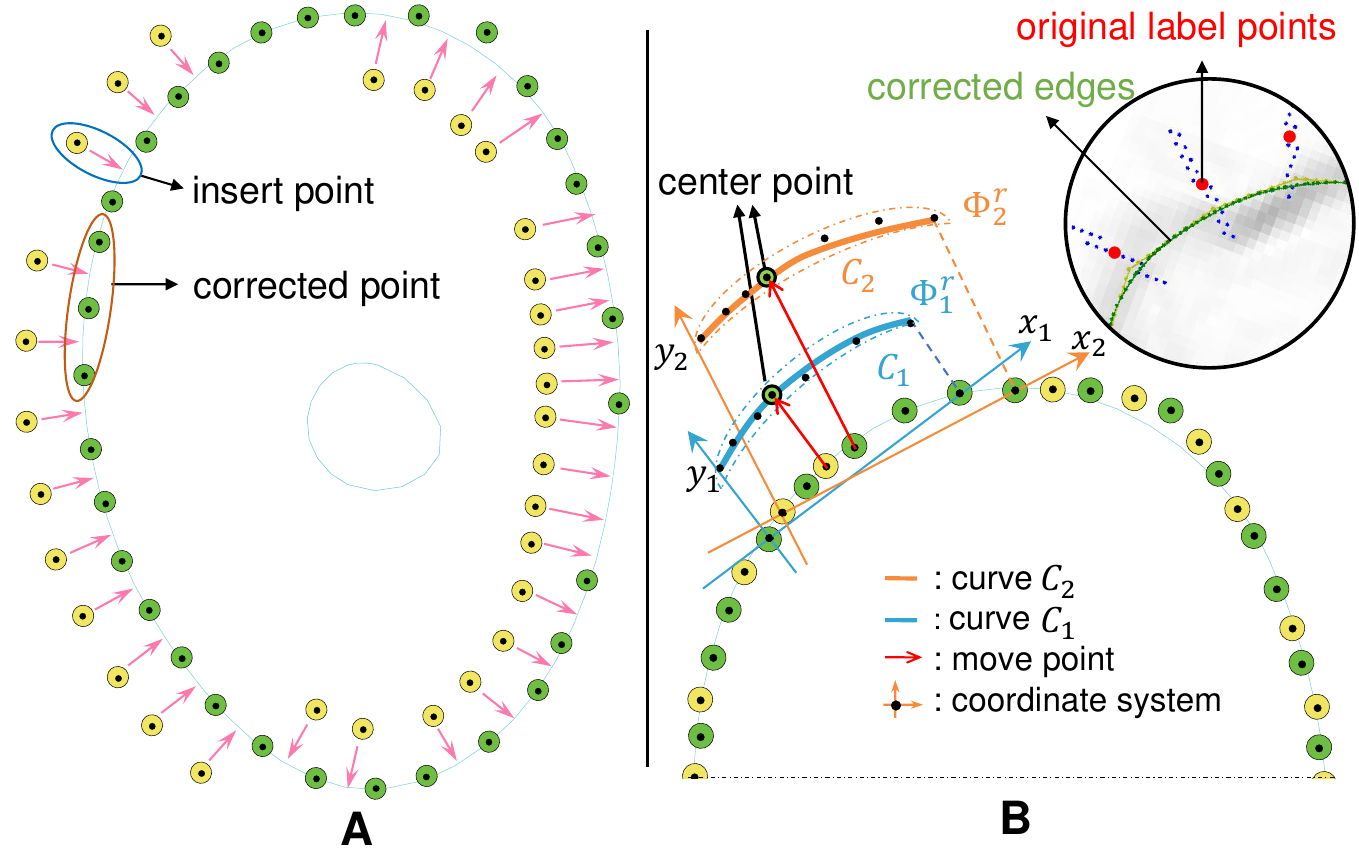}
  \caption{Local point smoothing to generate smooth closed discrete edges. {\bf (A)} Insert
  points; {\bf (B)} Move the coordinate system and correct each point by curve fitting.
  All corrected points constitute a discrete edge.}
  \label{fig4}
\end{figure}

\subsection{Local Point Smoothing} \label{section3c}
In Section~\ref{section3b}, we stitch multi-segment curves to obtain a closed cell curve,
and then sample the curve to generate a discrete edge.
In fact, there is no smoothness at the splice nodes.
To generate a smooth closed discrete edge,
we design a local point smoothing (LPS) method without curves splicing and sampling.
As shown in \textbf{\reffig{fig4}A}, we insert more
points in large intervals ($gap=1$).
As shown in \textbf{\reffig{fig4}B}, we only correct the center point of $\Phi _{k}^r$ by fitting a curve ($C_k$).
By shifting the local coordinate system by one step ($s = 1$), each point in $\left \{x_{I}^{j}\right \}$ will be corrected by fitting a curve.
These correction points constitute a discrete edge $C_d$.
Because no curves are spliced, the generated edge is smooth at each point.
The pipeline of our LLPC is shown in Algorithm~\ref{Algorithm1}.

\IncMargin{1em}
\begin{algorithm}
  \SetKwData{Left}{left}\SetKwData{This}{this}\SetKwData{Up}{up}
  \SetKwFunction{GPC}{GPC}\SetKwFunction{PI}{PI}\SetKwFunction{LPS}{LPS}\SetKwFunction{Append}{Append}
  \SetKwInOut{Input}{Input}\SetKwInOut{Output}{Output}
  \Input{$I$, a RGB image. $F$, a annotation file of $I$ ($F$ contains $k$ point lists). $r =15$, $\lambda_t=4$, $n_g=14$.}
  \Output{The corrected discrete edge $C_d$.}
  \BlankLine
  $C_d$$\leftarrow$ []\;
  \For{$i\leftarrow 1$ \KwTo $k$}{
    $x_s$$\leftarrow$ F[i]\;
      $x_c$$\leftarrow$ \GPC{$x_s,I,r,\lambda_t$} (based on Eq.~\ref{eq1})\;
    $x_I$$\leftarrow$ \PI{$x_c,gap=1$} (based on Eq.~\ref{eqpi})\;
      $x_t$$\leftarrow$ \LPS{$x_I,n_g$} (based on Eq.~\ref{eq6})\;
    $C_d$$\leftarrow$ \Append($C_d,x_t$)
  }
  \caption{LLPC Label Correction Algorithm.}\label{Algorithm1}
\end{algorithm}\DecMargin{1em}

\subsection{Parameter Setting}
In Section~\ref{section3a}, we set the parameters $r =15$, $\lambda_t=4$ and $h_1=r/2$.
In Section~\ref{section3b}, we set $n_g=14$. $r_f=\left \lfloor (n_g-1)/2 \right \rfloor $. $h_2=r_f/2$.
When $gap = 1$ and $s = 1$, {\bf the Section~\ref{section3c} is a special case of the Section~\ref{section3b}}.
See Section~\ref{section5c} for more discussion of parameter selection.

\section{Experimental design}
\subsection{Data Aquisition and Processing}
We compare our CCEDD with other cervical
cytology datasets in \textbf{Table~\ref{tab1}}. Our dataset was collected from Liaoning
Cancer Hospital $\&$ Institute between 2016 and
2017. We capture digital images with a Nikon ELIPSE Ci slide scanner,
SmartV350D lens and a 3-megapixel digital camera. For patients with negative and
positive cervical cancer, the optical magnification is 100$\times$
and 400$\times$, respectively. All of the cases are anonymized. All
processes of our research (image acquisition and processing,
etc.) follow ethical principles. Our CCEDD dataset includes
686 cervical images with a size of 2048$\times$1536 pixels (\textbf{Table~\ref{tab2}}).
Six expert cytologists outline the closed contours of the
cytoplasm and nucleus in cervical cytological images by
an annotation software (labelme~\cite{labelme2016}).

\begin{table}
  \caption{Comparison with other cervical cytology datasets. For a fair comparison of the sizes of different datasets, we crop the images to 512 $\times $ 512, and our CCEDD is about ten times larger than other datasets.}
  \label{tab1}
  \begin{center}
    \setlength{\tabcolsep}{1.6mm}{
      \begin{tabular}{lcccc}
        \toprule
        Dataset                             & Image size                  & Dataset size & Dataset size(512 $\times $ 512) & Open             \\
        \midrule
        ISBI~\cite{lu2015improved}          & 1024 $\times $ 1024         & 17           & 68                              & $\surd$          \\
        SZU Dataset~\cite{song2016accurate} & 1360 $\times $ 1024         & 21           & 84                              & $\times$         \\
        BHU Dataset~\cite{wan2019accurate}  & 512 $\times $ 512           & 580          & 580                             & $\times$         \\
        CCEDD                               & \textbf{2048 $\times $1536} & \textbf{686} & \textbf{8232}                   & \textbf{$\surd$} \\
        \bottomrule
      \end{tabular}
    }
  \end{center}
\end{table}

\begin{table}
  \caption{The detailed description of CCEDD.}
  \label{tab2}
  \begin{center}
    \setlength{\tabcolsep}{5.5mm}{
      \begin{tabular}{lcc}
        \toprule
        Our CCEDD           & Uncut CCEDD        & Cut CCEDD         \\
        \midrule
        Image size          & 2048 $\times $1536 & 512 $\times $ 384 \\
        Training set size   & 411                & 20139             \\
        Validation set size & 68                 & 3332              \\
        Test set size       & 207                & 10143             \\
        Dataset size        & 686                & 33614             \\
        \bottomrule
      \end{tabular}
    }
  \end{center}
\end{table}

\begin{figure}
  \centering
  \includegraphics[scale=0.35]{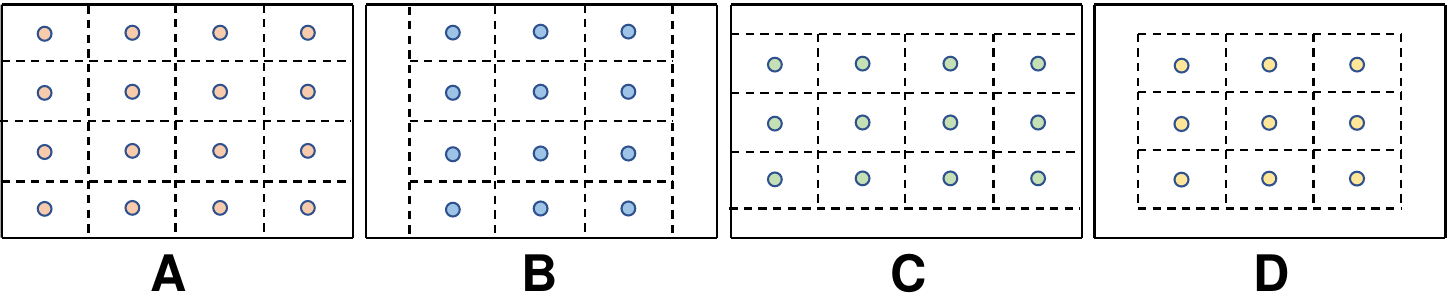}
  \caption{Image cutting method. {\bf (A)} 4$\times $4 cutting grid; {\bf (B)} move the grid right; {\bf (C)} move the grid down; {\bf (D)} move the grid right and down.}
  \label{fig5}
\end{figure}

We randomly shuffle our dataset and split it into training, validation and test sets.
To ensure test reliability, we set this ratio to 6:1:3.
To be able to train various complex neural networks on a GPU,
we crop a large-size image into small-size images.
If an image is cut as shown in \textbf{\reffig{fig5}A},
it will result in incomplete edge at the cut boundary.
To maximize data utilization efficiency,
we move the cutting grid, as shown in \textbf{\reffig{fig5}B, C, D}.
After label correction, we cut an image with
a size of 2048$\times$1536 into 49 image patches with a size of 512$\times$384 pixels.

\subsection{Baseline Model and Evaluation Metrics} \label{section4b}
\textbf{Baseline Model.}
Our baseline detectors are 10 state-of-the-art models.
We evaluate multiple edge detectors, such
as RCF~\cite{liuyun2019}, ENDE~\cite{Nazeri_2019_ICCV}, DexiNed~\cite{poma2020dense}, FINED~\cite{wibisono2020fined} and
PiDiNet~\cite{su2021pixel}.
Furthermore, we explore more network structures for edge detection by introducing segmentation networks, which usually only requires simple modifications of the last layer of networks.
These segmentation networks include STDC~\cite{Fan2021a}, UNet~\cite{ronneberger2015u}, UNet++~\cite{zhou2019unet++}, CENet~\cite{Gu2019}, MSU-Net~\cite{Su2021}.
To aggregate more shallow features for edge detection,
we modify multiple layers of STDC, i.e., STDC+.
More details of these network structure can be found in our code implementation.

\textbf{Evaluation Metrics.} We quantitatively evaluate the edge
detection accuracy by calculating three standard measures
(ODS, OIS and AP)~\cite{arbelaez2010contour}. The average precision (AP) is the
area under the precision-recall curve (\textbf{\reffig{fig1}B}).
F1-score$=\frac{2\cdot precision\cdot recall}{precision+ recall}$ is the harmonic average of precision
and recall. ODS is the best F1-score for a fixed scale, while OIS is
the F1-score for the best scale in each image.

\begin{figure*}
  \begin{center}
    \includegraphics[width=1\linewidth]{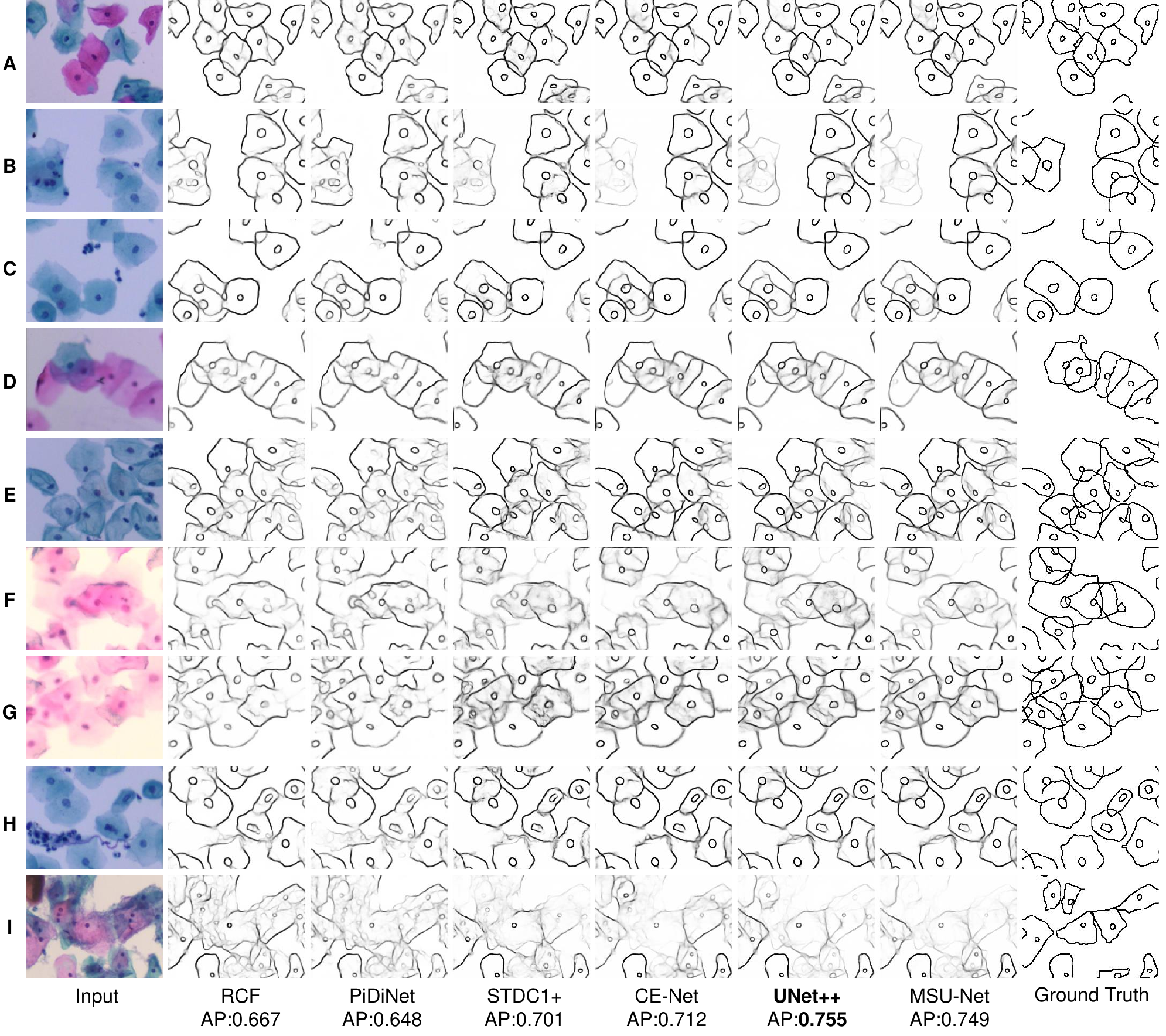}
  \end{center}
  \caption{Visual comparison results on CCEDD dataset. {\bf (A)} Slightly overlapping cells; {\bf (B)}{\bf (C)} Highly overlapping cells; {\bf (D)}{\bf (E)} Overlapping cell masses.
    {\bf (F)}{\bf (G)} Blurred overlapping cells; {\bf (H)}{\bf (I)} Overlapping cells in complex environments.}
  \label{fig6}
\end{figure*}

\subsection{Experimental Setup} \label{section4c}
\textbf{Training Strategy.} Data augmentation can improve model generalization and performance~\cite{bloice2019biomedical}.
In training, we perform rotation and shearing operations, which require padding zero pixels around an image.
In testing, there is no zero pixel padding.
This lead to different distributions of training and testing sets
and degrade the model performance.
Therefore, we perform data augmentation in pre-training and no augmentation during fine-tuning.

Due to the different structures and parameters of baseline networks,
a fixed number of training iterations may lead to overfitting or underfitting.
For accurate evaluation, we adaptively adjust the iteration number by evaluating the average accuracy (AP) on the validation set.
The period of model evaluation is set 1 epoch for pre-training and
0.1 epoch for fine-tuning. After the $i$-th model evaluation, we
can obtain $Model_i$ and $AP_i$ ($i=1,2,\cdots ,50$).
If $AP_i < min(AP_{i-j}$), the training ends and we obtain the
optimal model $Model_{j|max(AP_j)}$. $j = 1, 2, 3$ in pre-training
and $j = 1, 2, \cdots , 10$ in fine-tuning.
The maximum iteration number is 50 epochs for pre-training and fine-tuning.
Besides, we also dynamically adjust the learning rate to improve performance.
The learning rate $l$ decays from $1^{-4}$ to $1^{-5}$.
If $AP_i < AP_{i-1}$, $l_i = l_{i-1}/2$.

\textbf{Implementation Details.}
We use the Adam optimizer~\cite{kingma2015adam} to optimize all baseline networks on PyTorch ($\beta _1=0$, $\beta _2=0.9$).
We use random normal initialization to initialize these networks.
To be able to train various complex neural networks on a GPU, we resize the image to 256$\times $192.
The batch size is set 4.
We perform color adjustment, affine transformation and elastic
deformation for data augmentation~\cite{bloice2019biomedical}.
All experiments are implemented on a workstation equipped with a Intel Xeon Silver 4110 CPUs
and a NVIDIA RTX 3090 GPU.

\begin{table*}
  \caption{Edge detection results on our CCEDD dataset.
    Our baseline model contains RCF~\cite{liuyun2019}, ENDE~\cite{Nazeri_2019_ICCV}, DexiNed~\cite{poma2020dense},
    FINED~\cite{wibisono2020fined}, PiDiNet~\cite{su2021pixel}, STDC~\cite{Fan2021a}, UNet~\cite{ronneberger2015u},
    CE-Net~\cite{Gu2019}, UNet++~\cite{zhou2019unet++}, MSU-Net~\cite{Su2021}.
    "BCELoss" is binary cross entropy loss function.
    "RCFLoss" is an annotator-robust loss function for edge detection~\cite{liuyun2019}.
    STDC2~\cite{Fan2021a} has more parameters than STDC1~\cite{Fan2021a}.
    ”UNet++(DS)” is UNet++~\cite{zhou2019unet++} with deep supervision.
    ”MACs” is multiply–accumulate operation. ”Params” and ”MACs” are calculated by THOP\protect\footnotemark[1].
    Best and second best results are {\bf highlighted} and \underline{underlined}.}
  \label{tab4}
  \begin{center}
    \setlength{\tabcolsep}{0.5mm}{
      \begin{tabular}{l|c|c|c|c|c|c|c|c|c}
        \multirow{2}*{Year/Model/Loss} & \multirow{2}*{$\Delta $AP($\%$)} & \multicolumn{3}{c|}{Label correction} & \multicolumn{3}{c|}{No label correction} & \multirow{2}*{Params(M)} & \multirow{2}*{MACs{\bf (G)}}                                                                                                             \\
        \cline{3-8}                    &                                  & AP                                    & ODS                                      & OIS                      & AP                           & ODS               & OIS               &                                 &                                 \\
        \hline
        2019/RCF/RCFLoss               & 41.0$\%$                         & 0.612                                 & 0.599                                    & 0.594                    & 0.434                        & 0.485             & 0.485             & \multirow{2}*{14.81}            & \multirow{2}*{19.56}            \\
        2019/RCF/BCELoss               & \underline{41.9$\%$}             & 0.667                                 & 0.638                                    & 0.645                    & 0.470                        & 0.507             & 0.512             &                                 &                                 \\
        \hline
        2019/ENDE/BCELoss              & 37.0$\%$                         & 0.733                                 & 0.682                                    & 0.691                    & 0.535                        & \underline{0.548} & 0.555             & 6.06                            & 32.51                           \\
        \hline
        2020/DexiNed/RCFLoss           & 30.3$\%$                         & 0.649                                 & 0.633                                    & 0.635                    & 0.498                        & 0.528             & 0.533             & \multirow{2}*{35.08}            & \multirow{2}*{27.72}            \\
        2020/DexiNed/BCELoss           & 38.5$\%$                         & 0.723                                 & 0.671                                    & 0.680                    & 0.522                        & 0.541             & 0.549             &                                 &                                 \\
        \hline
        2020/FINED/RCFLoss             & 28.4$\%$                         & 0.602                                 & 0.604                                    & 0.450                    & 0.469                        & 0.510             & 0.402             & \multirow{2}*{\underline{1.43}} & \multirow{2}*{14.38}            \\
        2020/FINED/BCELoss             & 41.4$\%$                         & 0.703                                 & 0.660                                    & 0.621                    & 0.497                        & 0.528             & 0.530             &                                 &                                 \\
        \hline
        2021/PiDiNet/RCFLoss           & 37.2$\%$                         & 0.590                                 & 0.581                                    & 0.574                    & 0.430                        & 0.481             & 0.479             & \multirow{2}*{{\bf 0.69}}       & \multirow{2}*{{\bf 3.74}}       \\
        2021/PiDiNet/BCELoss           & {\bf 42.7$\%$}                   & 0.648                                 & 0.624                                    & 0.628                    & 0.454                        & 0.496             & 0.501             &                                 &                                 \\
        \hline
        \hline
        2021/STDC1/BCELoss             & 12.9$\%$                         & 0.394                                 & 0.466                                    & 0.472                    & 0.349                        & 0.438             & 0.443             & \multirow{2}*{14.26}            & \multirow{2}*{\underline{4.48}} \\
        2021/STDC1(pretrain)/BCELoss   & 13.1$\%$                         & 0.407                                 & 0.478                                    & 0.483                    & 0.360                        & 0.451             & 0.454             &                                 &                                 \\

        \hline
        2021/STDC2/BCELoss             & 16.1$\%$                         & 0.403                                 & 0.473                                    & 0.478                    & 0.347                        & 0.435             & 0.442             & \multirow{2}*{22.30}            & \multirow{2}*{7.01}             \\
        2021/STDC2(pretrain)/BCELoss   & 15.0$\%$                         & 0.413                                 & 0.484                                    & 0.488                    & 0.359                        & 0.449             & 0.454             &                                 &                                 \\
        \hline
        2021/STDC1+/BCELoss            & 41.3$\%$                         & 0.701                                 & 0.652                                    & 0.659                    & 0.496                        & 0.518             & 0.524             & 13.76                           & 39.28                           \\
        2021/STDC2+/BCELoss            & 38.2$\%$                         & 0.694                                 & 0.648                                    & 0.656                    & 0.502                        & 0.525             & 0.532             & 21.83                           & 41.81                           \\
        \hline
        \hline
        2015/UNet/BCELoss              & 38.9$\%$                         & 0.729                                 & 0.679                                    & 0.689                    & 0.525                        & 0.539             & 0.546             & 31.03                           & 41.96                           \\
        \hline
        2019/CE-Net(pretrain)/BCELoss  & 37.5$\%$                         & 0.696                                 & 0.653                                    & 0.658                    & 0.506                        & 0.530             & 0.535             & \multirow{2}*{60.24}            & \multirow{2}*{17.36}            \\
        2019/CE-Net/BCELoss            & 36.4$\%$                         & 0.712                                 & 0.668                                    & 0.675                    & 0.522                        & 0.540             & 0.547             &                                 &                                 \\
        \hline
        2019/UNet++(DS)/BCELoss        & 37.6$\%$                         & 0.739                                 & 0.687                                    & 0.696                    & \underline{0.537}            & \underline{0.548} & 0.555             & \multirow{2}*{9.16}             & 26.76                           \\
        2019/UNet++/BCELoss            & 39.6$\%$                         & {\bf 0.755}                           & {\bf 0.691}                              & {\bf 0.701}              & {\bf 0.541}                  & {\bf 0.550}       & {\bf 0.557}       &                                 & 26.75                           \\
        \hline
        2021/MSU-Net/BCELoss           & 39.7$\%$                         & \underline{0.749}                     & \underline{0.689}                        & \underline{0.699}        & 0.536                        & {\bf 0.550}       & \underline{0.556} & 47.09                           & 59.93                           \\
      \end{tabular}
    }
  \end{center}
\end{table*}
\begin{figure}
  \begin{center}
    \includegraphics[width=0.6\linewidth]{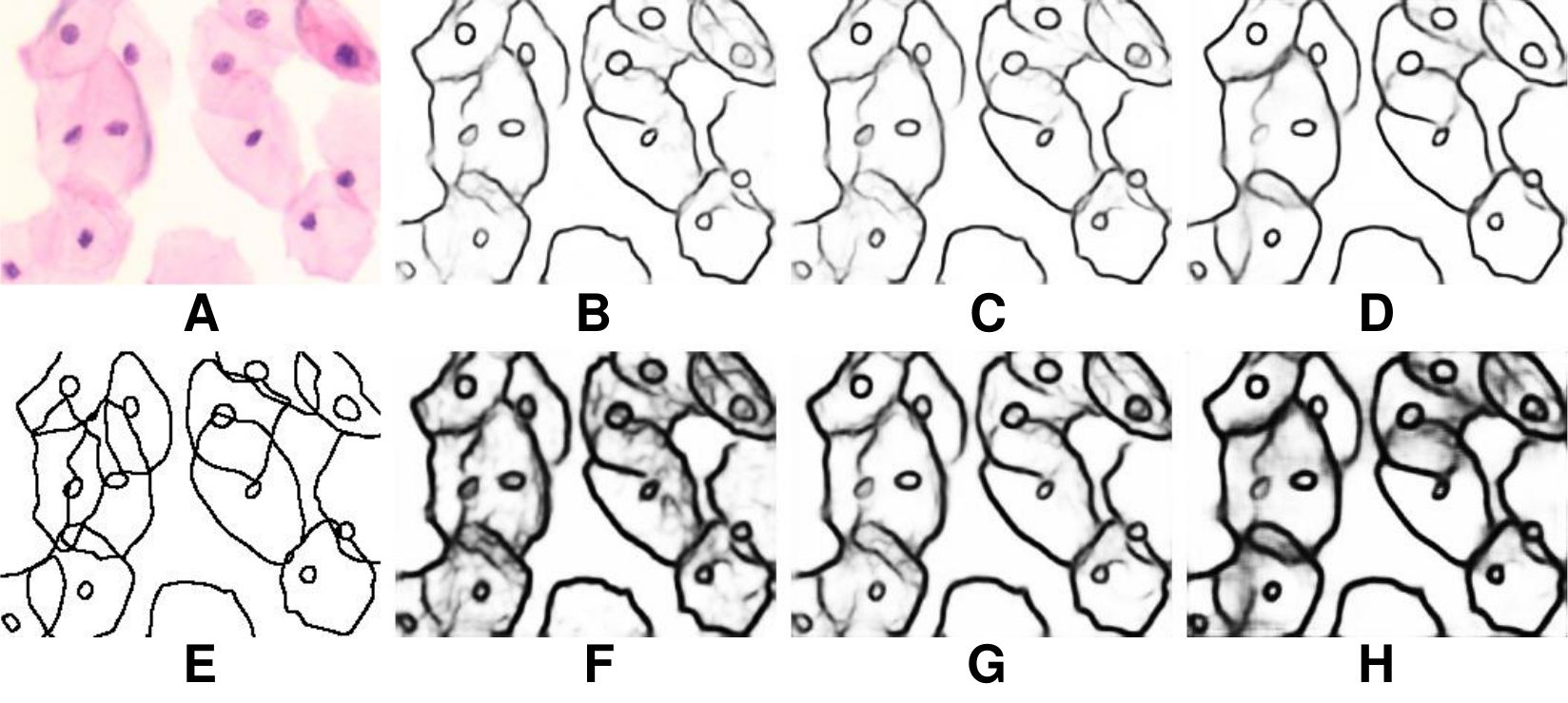}
  \end{center}
  \caption{Visual comparison of different loss functions.
    {\bf (A)} Input; {\bf (E)} Ground Truth; {\bf (B)}{\bf (F)} PiDiNet~\cite{su2021pixel}; {\bf (C)}{\bf (G)} RCF~\cite{liuyun2019}; {\bf (D)}{\bf (H)} DexiNed~\cite{poma2020dense};
  {\bf (B)}{\bf (C)}{\bf (D)} BCELoss; {\bf (F)}{\bf (G)}{\bf (H)} RCFLoss~\cite{liuyun2019}.
  "BCELoss" is binary cross entropy loss function.
  Compared with BCELoss, RCFLoss~\cite{liuyun2019} can produce coarser edges.}
  \label{fig8}
\end{figure}
\section{Experimental Results and Discussion}
\subsection{Edge Detection of Overlapping Cervical Cells}\label{section5a}
We show the visual comparison results on our CCEDD in \textbf{\reffig{fig6}}. The quantitative comparison is shown in
\textbf{Table~\ref{tab4}} and \textbf{\reffig{fig1}B}. These results have important guiding
implications for accurate edge detection of overlapping cervical cells.
We analyze several factors affecting the performance of overlapping edge detection.

\begin{itemize}
  \item Loss function design. RCFLoss~\cite{liuyun2019} produces coarser
        edges, as shown in \textbf{\reffig{fig8}}. This may be robust for natural images, but poor localization accuracy for accurate cervical cell edge detection.
  \item Network structure design.
        Long-distance skip connections can fuse shallow and deep features for constructing multi-scale features.
        Our experiments show that the U-shaped structure is effective for overlapping edge detection (e.g., UNet~\cite{ronneberger2015u}, UNet++~\cite{zhou2019unet++} and MSU-Net~\cite{Su2021}).
  \item Pre-training.
        Due to the huge distribution difference between natural and medical images,
        pre-training may degrade performance (e.g., CE-Net~\cite{Gu2019}) or have limited improvement (e.g., STDC~\cite{Fan2021a}).
\end{itemize}

\footnotetext[1]{https://github.com/Lyken17/pytorch-OpCounter.}

\begin{figure}
  \centering
  \includegraphics[width=0.6\linewidth]{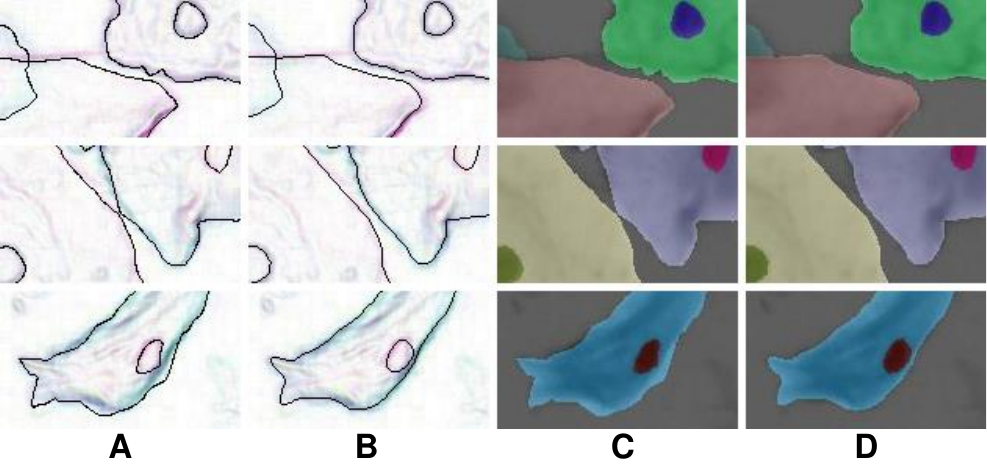}
  \caption{Label correction for edge detection and
  semantic segmentation. {\bf (A)} Original edge; {\bf (B)} Corrected edge;
  {\bf (C)} Original mask; {\bf (D)} Corrected mask.}
  \label{fig11}
\end{figure}
\begin{table}
  \caption{Performance improvement analysis of label correction.
    Use UNet++~\cite{zhou2019unet++} for evaluation.}
  \label{tab5}
  \begin{center}
    \setlength{\tabcolsep}{5.5mm}{
      \begin{tabular}{lccc}
        \toprule
        Training / Evaluation             & AP          & ODS            & OIS            \\
        \midrule
        Original label / Original label   & 0.541       & 0.550          & 0.557          \\
        Original label / Corrected label  & 0.588       & 0.592          & 0.598          \\
        Corrected label / Corrected label & {\bf 0.755} & \textbf{0.691} & \textbf{0.701} \\
        \bottomrule
      \end{tabular}
    }
  \end{center}
\end{table}

\subsection{Effectiveness of Label Correction}
In our LLPC, the position of label points is locally corrected to the pixel gradient peak.
As shown in \textbf{\reffig{fig1}A} and \textbf{\reffig{fig11}B}, Our LLPC can generate more accurate edge labels.
Besides, we can easily generate corrected masks from corrected points in the labelme software~\cite{labelme2016}.
Compared with the original mask in \textbf{\reffig{fig11}C}, our corrected mask has higher edge localization accuracy and smoother edges, as shown in \textbf{\reffig{fig11}D}.

\begin{figure*}
  \centering
  \includegraphics[width=1\linewidth]{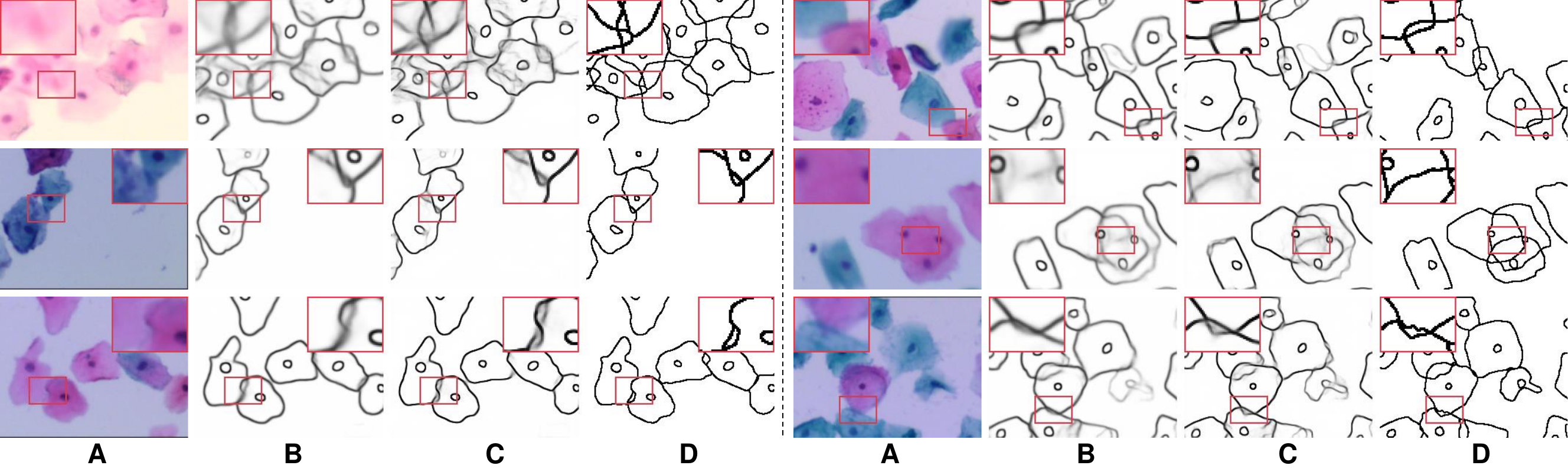}
  \caption{Visual comparison results of training with different labels. {\bf (A)} Input image; {\bf (B)} UNet++~\cite{zhou2019unet++}/BCELoss + Original
  label; {\bf (C)} UNet++~\cite{zhou2019unet++}/BCELoss + Corrected label; {\bf (D)} Corrected
  labels. Compared with the original label, the corrected
  label can improve the accuracy of overlapping
  edge detection.}
  \label{fig7}
\end{figure*}

We train multiple networks using original label and corrected label.
The quantitative comparison results is shown in \textbf{Table~\ref{tab4}} and \textbf{\reffig{fig1}B}. Compared with the original
label, using the corrected label to train multiple networks
can significantly improve AP ({\bf 30-40$\%$}), which verifies the effectiveness of our label correction method.
\textbf{Table~\ref{tab5}} shows that
the performance improvement comes from two aspects.
First, our corrected label can improve the evaluation accuracy in testing (0.541$\to $0.588).
Second, using our corrected label to train network can improve the accuracy of overlapping
edge detection in training (0.588$\to $0.755), as shown in \textbf{\reffig{fig7}}.

\begin{figure}
  \centering
  \includegraphics[scale=0.45]{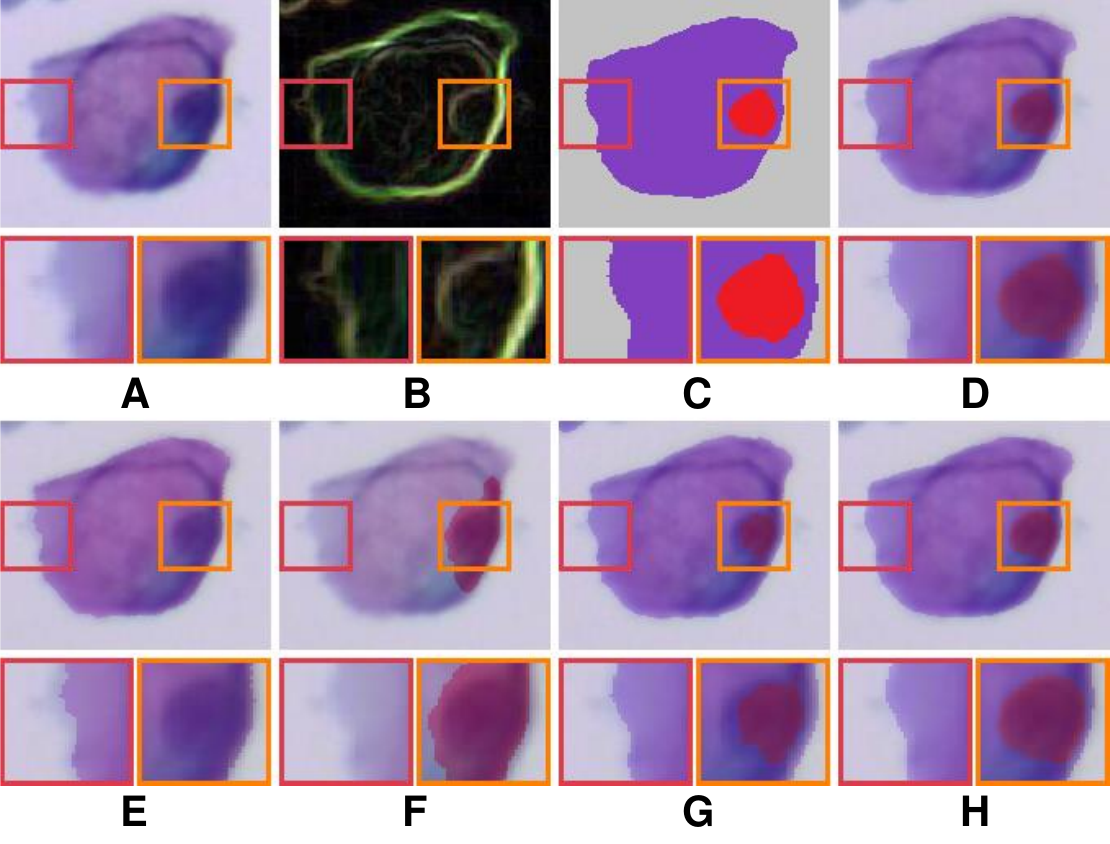}
  \caption{Qualitative comparison of single-cell label correction.
    {\bf (A)} Input; {\bf (B)} Gradient image; {\bf (C)} Original mask; {\bf (D)} Input + original mask;
  {\bf (E)} Active contours~\cite{Chan2001} for cytoplasm; {\bf (F)} Active contours~\cite{Chan2001} for nucleus; {\bf (G)} Dense CRF~\cite{Krahenbuhl2011}; {\bf (H)} Our LLPC.}
  \label{fig18}
\end{figure}

\subsection{Comparison with Other Label Correction Methods}\label{section5c_crf}
\begin{figure}
  \centering
  \includegraphics[width=0.6\linewidth]{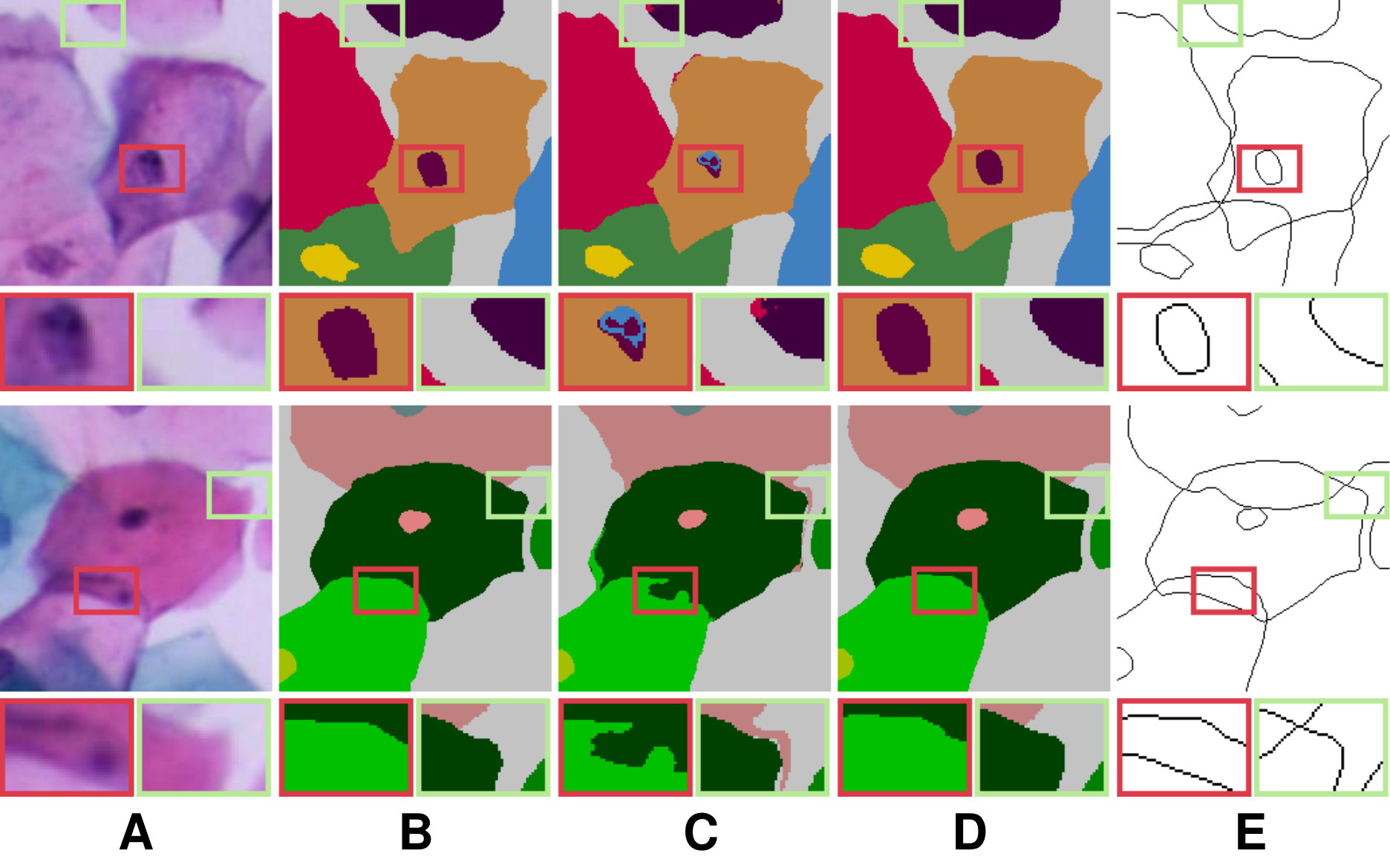}
  \caption{Qualitative comparison of label correction for overlapping cell masses.
    {\bf (A)} Input; {\bf (B)} Original mask;
  {\bf (C)} Dense CRF~\cite{Krahenbuhl2011}; {\bf (D)} Our LLPC (mask); {\bf (E)} Our LLPC (edge).}
  \label{fig19}
\end{figure}

In \textbf{\reffig{fig18}} and \textbf{\reffig{fig19}}, we compare our LLPC with active contours~\cite{Chan2001} and dense CRF~\cite{Krahenbuhl2011}.
We observed that active contours~\cite{Chan2001} is refinement failure of nucleus contours in \textbf{\reffig{fig18}F},
and dense CRF~\cite{Krahenbuhl2011} fails due to complex overlapping cell contours in \textbf{\reffig{fig19}C}.
Since active contours~\cite{Chan2001} and dense CRF~\cite{Krahenbuhl2011} are global iterative optimization methods based on segmented images,
which are uncontrollable for label correction of object contours and ultimately lead to these failed results.
Our LLPC is the local label point correction without iterative optimization.
Therefore, the correction error of our LLPC is controllable
and the error in one place does not spread to other places, which is crucial for robust label correction.
Besides, dense CRF~\cite{Krahenbuhl2011} is nonplussed over overlapping instance segmentation refinement,
while our LLPC corrects label based on annotation point and can handle overlapping label correction, as shown in \textbf{\reffig{fig19}E}.

\begin{figure}
  \centering
  \includegraphics[width=0.55\linewidth]{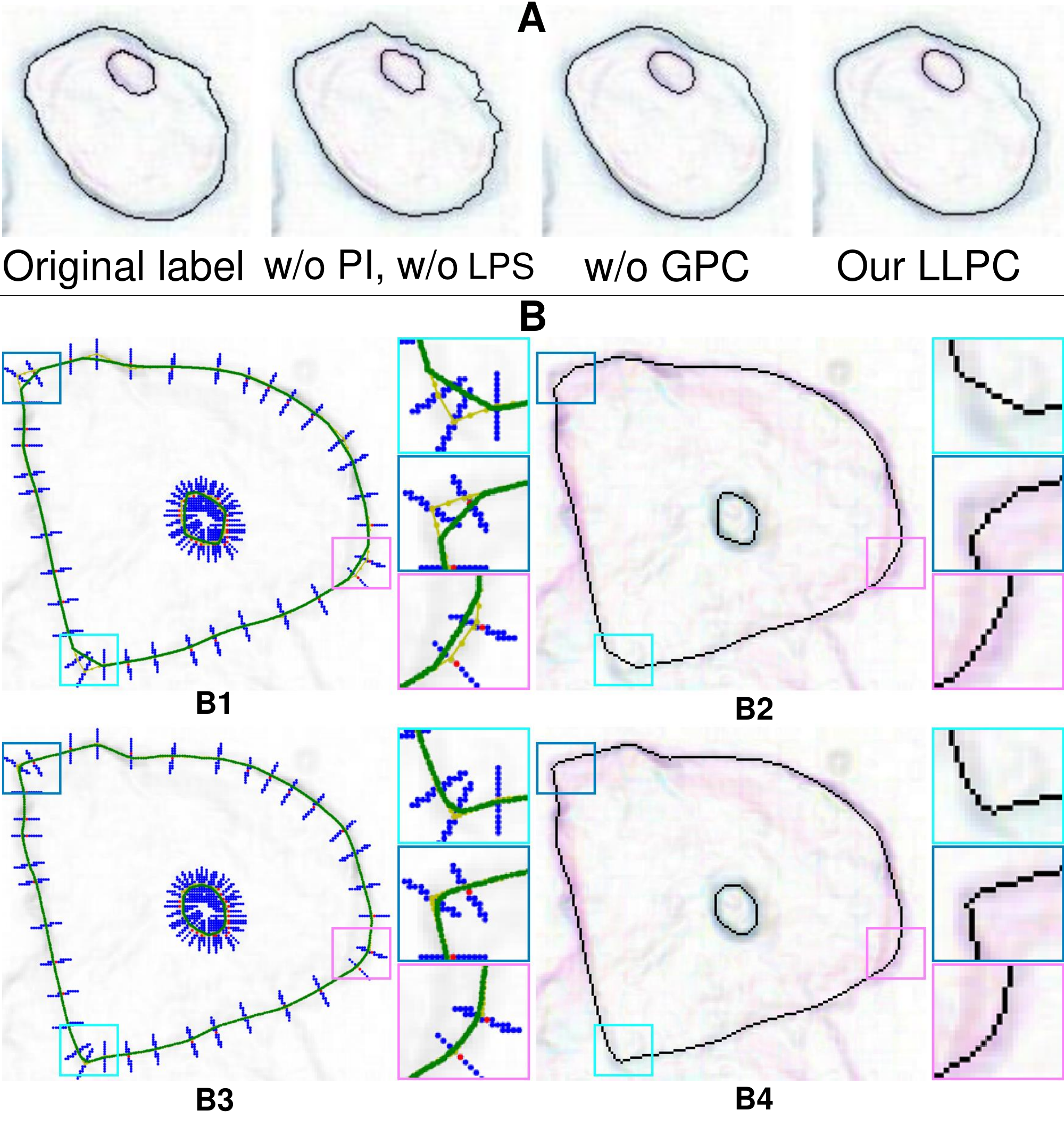}
  \caption{
    {\bf (A)} Ablation of gradient-guided point correction. "GPC" is gradient-guided point correction.
      {\bf (B)} Visual comparison of different label correction methods.
    (B1) The green curves generated by piecewise curve fitting (w/o LPS);
    (B2) The discrete edge sampled from the curves in (B1);
    (B3) The curves smoothed by LPS; (B4) The discrete edges without curve sampling.}
  \label{figcorrect}
\end{figure}

\begin{table}
  \caption{Ablation of our LLPC. "GPC" is gradient-guided point correction. "PI" is point interpolation.
    "w/o LPS" is using piecewise curve fitting instead of local point smoothing. Use UNet++~\cite{zhou2019unet++} for evaluation.}
  \label{tabAblationLabelCorrection}
  \begin{center}
    \setlength{\tabcolsep}{5.5mm}{
      \begin{tabular}{lccc}
        \toprule
        Correction Method     & AP          & ODS            & OIS            \\
        \midrule
        Original label        & 0.541       & 0.550          & 0.557          \\
        GPC (w/o PI, w/o LPS) & 0.731       & 0.682          & 0.692          \\
        {\bf Our LLPC}        & {\bf 0.755} & \textbf{0.691} & \textbf{0.701} \\
        \hline
        w/o GPC               & 0.533       & 0.545          & 0.552          \\
        w/o PI                & 0.663       & 0.619          & 0.625          \\
        w/o LPS               & 0.742       & 0.689          & 0.699          \\
        \hline

        \bottomrule
      \end{tabular}
    }
  \end{center}
\end{table}

\subsection{Ablation experiment} \label{section5c}
\textbf{Ablation of Label Correction Method.}
Our LLPC contains three steps: gradient-guided point correction (GPC), point interpolation (PI) and local point smoothing (LPS).
Although our GPC can correct label points to pixel gradient peaks,
there is still some error in the correction process.
LPS can smooth the edges corrected by GPC, as shown in \textbf{\reffig{figcorrect}A}.
\textbf{Table~\ref{tabAblationLabelCorrection}} shows that GPC is the most important part of our LLPC (0.541$\to$0.731),
while PI and LPS can further improve the annotation quality by smoothing edges (0.731$\to$0.755).
Only smoothing the original labels ("w/o GPC") is ineffective (0.541$\to$0.533).
Because this may lead to larger annotation errors.
Compared to piecewise curve fitting in Section~\ref{section3b},
LPS can generate smoother edges, as shown in \textbf{\reffig{figcorrect}B}.
These qualitative and quantitative results verify that the three components of our LLPC are essential.
\begin{table}
  \caption{Parameters ablation of our label correction method. Use UNet++~\cite{zhou2019unet++} for evaluation.
    For our CCEDD, we set $r=15, \lambda _t=4$ and $n_{g}=14$.}
  \label{tab6}
  \begin{center}
    \setlength{\tabcolsep}{3.8mm}{
      \begin{tabular}{ccccccc}
        \toprule
        $r$ & $\lambda _t$ & $gap$ & $n_{g}$ & AP          & ODS            & OIS            \\
        \midrule
        7   & 4            & 1     & 14      & 0.691       & 0.645          & 0.653          \\
        11  & 4            & 1     & 14      & 0.732       & 0.681          & 0.691          \\
        19  & 4            & 1     & 14      & 0.746       & 0.691          & 0.701          \\
        23  & 4            & 1     & 14      & 0.734       & 0.683          & 0.692          \\
        \hline
        15  & 1            & 1     & 14      & 0.750       & 0.689          & 0.700          \\
        15  & 2            & 1     & 14      & 0.751       & 0.690          & 0.700          \\
        15  & 3            & 1     & 14      & 0.745       & 0.691          & 0.700          \\
        15  & 5            & 1     & 14      & 0.750       & 0.689          & 0.699          \\
        15  & 10           & 1     & 14      & 0.729       & 0.679          & 0.688          \\
        15  & 15           & 1     & 14      & 0.708       & 0.658          & 0.664          \\
        \hline

        15  & 4            & 1.5   & 14      & 0.749       & 0.688          & 0.698          \\
        15  & 4            & 2     & 14      & 0.742       & 0.689          & 0.699          \\
        \hline
        15  & 4            & 1     & 10      & 0.750       & 0.689          & 0.699          \\
        15  & 4            & 1     & 12      & 0.729       & 0.687          & 0.697          \\
        15  & 4            & 1     & 16      & 0.752       & 0.690          & 0.700          \\
        15  & 4            & 1     & 18      & 0.750       & 0.687          & 0.698          \\
        \hline
        15  & 4            & 1     & 14      & {\bf 0.755} & \textbf{0.691} & \textbf{0.701} \\
        \bottomrule
      \end{tabular}
    }
  \end{center}
\end{table}

\begin{figure}
  \centering
  \includegraphics[scale=0.5]{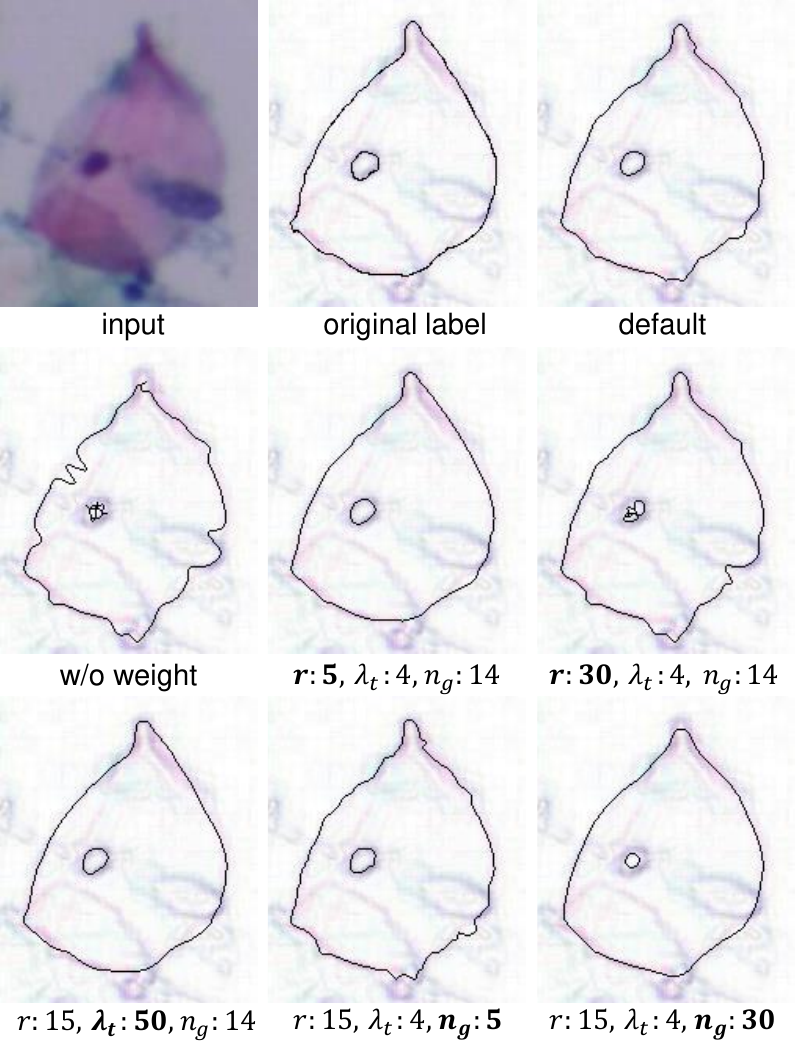}
  \caption{Visual comparison results for different parameters settings in our LLPC.
    "default" is $r=15, \lambda _t=4$ and $n_{g}=14$.
    "w/o weight" is $\omega_{j}=1$ in Eq.~\ref{eq3}.}
  \label{fig14}
\end{figure}

\textbf{Selection of hyper-parameters.}
To set the optimal parameters, we conduct parameters ablation experiments in \textbf{Table~\ref{tab6}}.
$gap$ can control the point density in PI. For local curve fitting, $gap=1$ is optimal.
Therefore, \textbf{for an unknown dataset, our LLPC only needs to set three parameters}, i.e., $r, \lambda _t$ and $n_{g}$.
A qualitative comparison of these parameters with different settings is shown in \textbf{\reffig{fig14}}.
$r$ controls the maximum error correction range in human annotations.
If $r$ is too small, large label errors cannot be corrected. If $r$ is too large, the error of point correction is larger.
$r$ limits the correction range in space,
while $\lambda _t$ is the threshold for a limitation of gradient values variation during the correction process.
If $\lambda_t$ is large, label points are corrected only when the gradient value changes sharply in the search direction.
$n_{g}$ controls the scale of the local smoothing.
For our CCEDD, $r=15, \lambda _t=4$ and $n_{g}=14$.

\begin{table}
  \caption{Ablation of two-stage training strategy.
    We perform data augmentation in pre-training and no augmentation during fine-tuning.
    Use UNet++~\cite{zhou2019unet++} for evaluation.}
  \label{tab3}
  \begin{center}
    \setlength{\tabcolsep}{3.7mm}{
      \begin{tabular}{lccc}
        \toprule
        Training methods                  & AP          & ODS         & OIS         \\
        \midrule
        w/o augmentation, w/o fine-tuning & 0.729       & 0.672       & 0.683       \\
        w/ augmentation, w/o fine-tuning  & 0.732       & 0.674       & 0.682       \\
        w/ augmentation, w/ fine-tuning   & {\bf 0.755} & {\bf 0.691} & {\bf 0.701} \\
        \bottomrule
      \end{tabular}
    }
  \end{center}
\end{table}

\begin{figure*}
  \centering
  \includegraphics[width=1\linewidth]{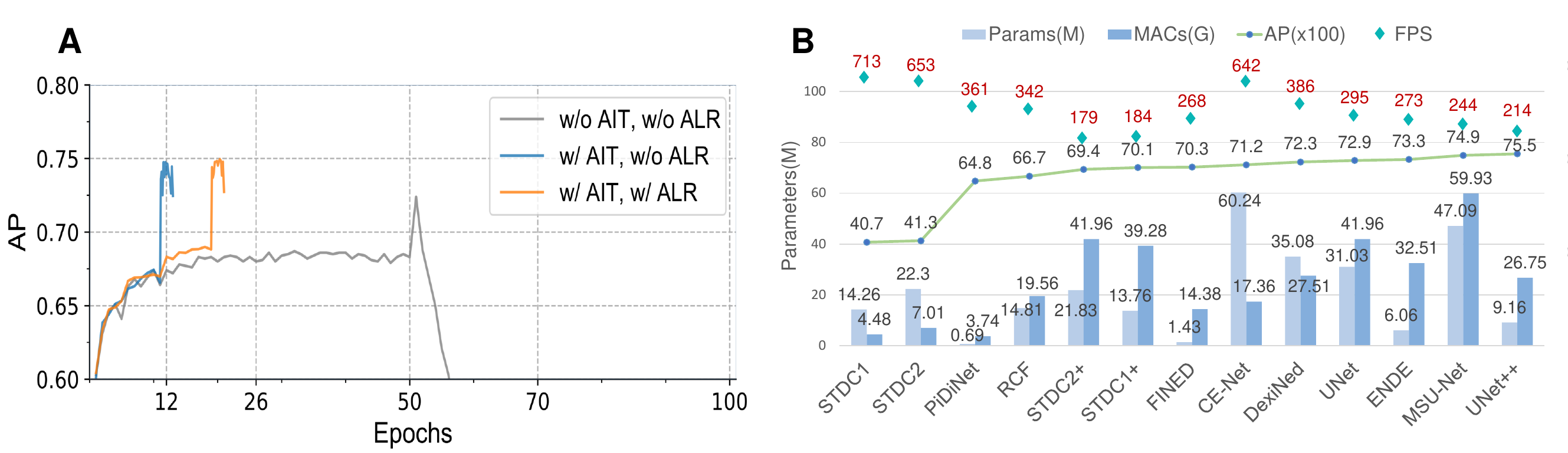}
  \caption{{\bf (A)} Training schedules.
    "AIT" is adaptive iteration training. "ALR" is adaptive learning rate.
    We evaluate UNet++~\cite{zhou2019unet++} on the validation set.
      {\bf (B)} Comparison of network parameters, running efficiency and edge detection performance. ”MACs” is multiply–accumulate operation.
    "FPS" is the average speed by evaluating 10413 images with a resolution of 256$\times $192.}
  \label{fig15}
\end{figure*}

\textbf{Ablation of Training Strategy.} Our training strategy can
eliminate the influence of different distributions of the training
and test sets due to data augmentation, and improve the AP by 3.6$\%$ in \textbf{Table~\ref{tab3}}.
To fairly evaluate multiple networks with different structures and parameters,
we employ adaptive iteration and learning rate adjustment to avoid overfitting and underfitting.
\textbf{Table~\ref{tab3-1}} and \textbf{\reffig{fig15}A} verify the effectiveness of our adaptive training strategy.
\begin{table}
  \caption{Ablation of adaptive training strategy. We evaluate UNet++~\cite{zhou2019unet++} on the validation set.
    "AIT" is adaptive iteration training. "ALR" is adaptive learning rate.}
  \label{tab3-1}
  \begin{center}
    \setlength{\tabcolsep}{4mm}{
      \begin{tabular}{lcccc}
        \toprule
        Training methods & AP          & ODS         & OIS         & epoch    \\
        \midrule
        w/o AIT, w/o ALR & 0.683       & 0.639       & 0.642       & 50       \\
        w/o AIT, w/o ALR & 0.449       & 0.653       & 0.657       & 70       \\
        w/o AIT, w/o ALR & 0.308       & 0.647       & 0.653       & 100      \\
        w/ AIT, w/o ALR  & 0.747       & 0.684       & 0.693       & {\bf 13} \\
        w/ AIT, w/ ALR   & {\bf 0.750} & {\bf 0.693} & {\bf 0.700} & 21       \\
        \bottomrule
      \end{tabular}
    }
  \end{center}
\end{table}

\subsection{Computational Complexity}
\textbf{Label Correction.} Our LLPC takes 270 seconds to generate 100 corrected edge images with a size of
2048$\times $1536 pixels on CPU. Because our label
correction algorithm is offline and does not affect the inference
time of a neural network, we have not further optimized it. If
the algorithm runs on GPU, the speed can be further improved,
which can save more time for label correction of large-scale
datasets.

\textbf{Model Evaluation.}
We rewrite the evaluation code~\cite{arbelaez2010contour}
on GPU for fast evaluation. The average FPS using the
UNet++~\cite{zhou2019unet++} is 173 for 10143 test images with a size of
256$\times $192 pixels. In training, we need to calculate the AP of the
validation set to adaptively control the learning rate and the
number of iterations (see Section~\ref{section4c}). Fast evaluation greatly
accelerates our training process.

\textbf{Neural Network Inference.}
We test the inference speed of UNet++~\cite{zhou2019unet++}.
For 207 images with a resolution of
1024$\times $768, the average FPS is 9. For 207 images with a
resolution of 512$\times $512, the average FPS is 26. For 10413
images with a resolution of 256$\times $192, the average FPS is 295.
\textbf{\reffig{fig15}B} shows the running efficiency comparison of multiple benchmark models.
According to the report of Wan \textit{et al.}~\cite{wan2019accurate},
the methods of~\cite{wan2019accurate,lu2015improved,lu2016evaluation},
took 17.67s, 35.69s and 213.62s for an image
a resolution of 512$\times $512, respectively. Compared with these
method, the UNet++~\cite{zhou2019unet++} is significantly faster. Many cervical cell
segmentation approaches~\cite{wan2019accurate,zhang2020polar,tareef2017automatic,phoulady2017framework,tareef2018multi} consist
of three stages, including nucleus candidate detection, cell
localizations and cytoplasm segmentation. Fast edge detection of
overlapping cervical cell means that
the detected edges can be used as a priori input of these
segmentation networks to improve performance at a small cost.

\begin{figure}
  \centering
  \includegraphics[scale=0.4]{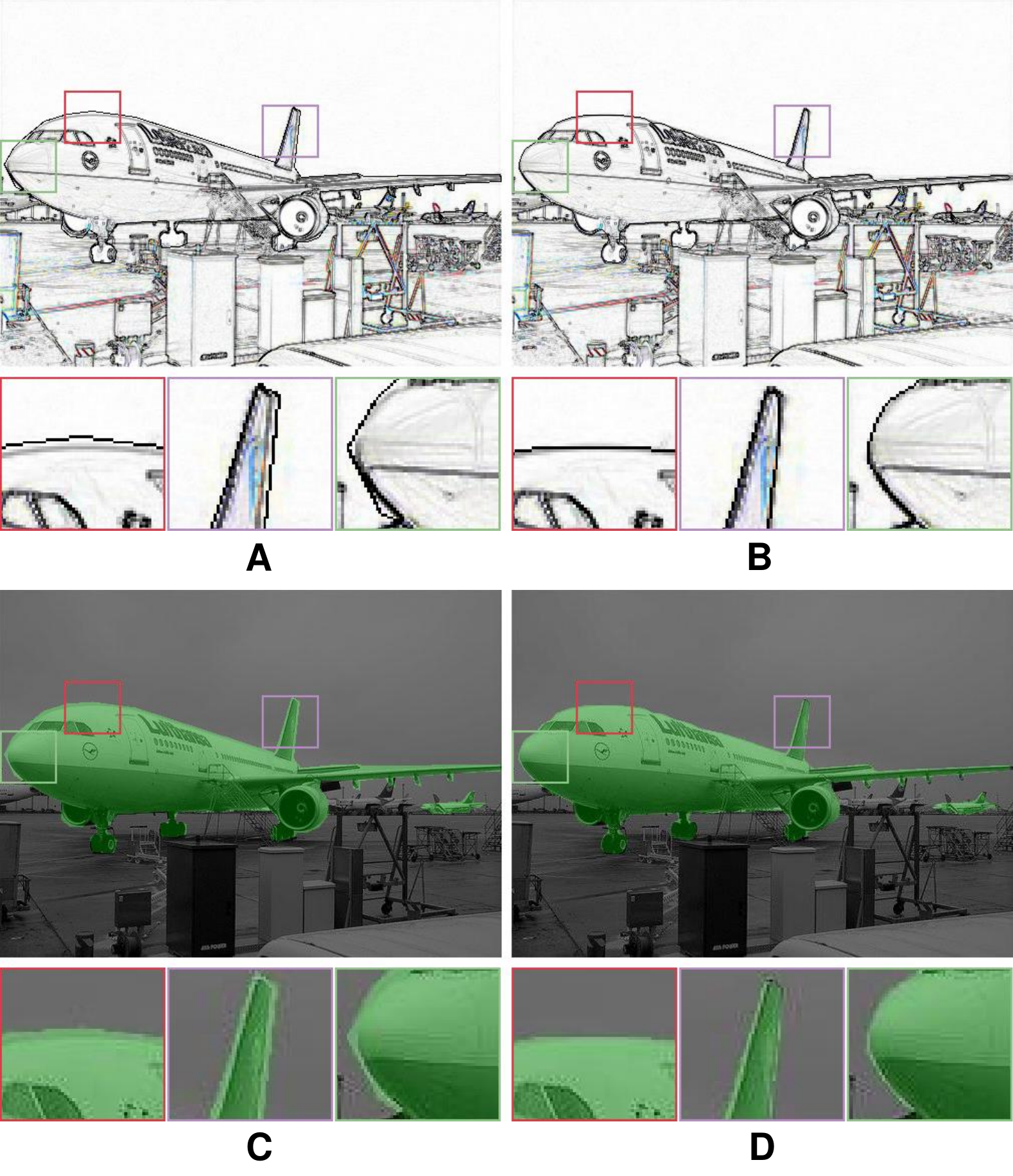}
  \caption{Label correction for natural images. {\bf (A)} Original edge; {\bf (B)} Corrected edge;
  {\bf (C)} Original mask; {\bf (D)} Corrected mask.}
  \label{fig12}
\end{figure}

\section{Discussion}
\textbf{Label Correction for Natural Images.}
Our label correction method can correct a closed contour by correcting the position of label points,
which does not require additional prior assumptions (e.g., contour shape, object size).
We annotated several images in the PASCAL VOC dataset~\cite{everingham2010pascal}
with labelme~\cite{labelme2016} and corrected the label ($r=7, \lambda _t=4$ and $n_{g}=9$).
As shown in \textbf{\reffig{fig12}}, our label correction method can generate more accurate object contours, which demonstrates the feasibility of our label correction method for natural images.

\textbf{Overlapping Edge Detection.}
Overlapping edge detection of cervical cell is a challenging task due to the presence of strong and weak gradient edges.
For edges with strong gradients, it only requires low-level detail features.
For edges with weak gradients in overlapping region, it may require high-level semantics to reason contours and connect edges
based on the context in strong gradient regions.
While Unet++~\cite{zhou2019unet++} achieves the best results on our CCEDD,
there is no difference in the detection of these two different types of edges.
Designing new network structures and loss functions for overlapping edge detection may be a way to further address this challenge.

\section{Conclusions}
We propose a local label point correction method for edge detection
and image segmentation, which is the first benchmark for label correction based on annotation points.
Our LLPC can improve the edge localization accuracy and mitigate labeling error from different annotators in
manual annotation.
Only 3 parameters need to be set in our LLPC, but using the
label corrected by our LLPC to train multiple networks can yield 30-40$\%$ AP improvement.
Besides, we construct a largest overlapping cervical cell edge detection dataset based on our LLPC,
which will greatly facilitate the development of overlapping cell edge detection.
In future work, we plan to develop a label point correction method with local adaptive parameter adjustment.

\section*{Conflict of Interest Statement}

The authors confirm that there are no conflicts of interest.

\section*{Author Contributions}
JL: conceptualization, methodology, software, validation, writing - original draft, visualization.
HF: investigation, resources, writing - review $\&$ editing, supervision, project administration, funding acquisition.
QW: writing - review $\&$ editing.
WL: investigation.
YT: writing - review $\&$ editing, supervision.
DW: investigation, resources, data curation.
MZ and LC: investigation, resources.
All authors contributed to the article and approved the
submitted version.

\section*{FUNDING}
This work is supported by the National Natural Science Foundation of China (61873259, 62073205, 61821005),
the Key Research and Development Program of Liaoning (2018225037),
and the Youth Innovation Promotion Association of Chinese Academy of Sciences (2019203).


\section*{Data Availability Statement}
The datasets for this study can be found in https://github.com/nachifur/LLPC.

{\small
\bibliographystyle{IEEEtran}
\bibliography{egbib}
}

\end{document}